\def\year{2019}\relax
\documentclass[letterpaper]{article} 
\usepackage{aaai19}  
\usepackage{times}  
\usepackage{helvet}  
\usepackage{courier}  
\usepackage{url}  
\usepackage{graphicx}  

\usepackage{amsmath}
\usepackage{amssymb}
\usepackage{tabularx}
\usepackage{multirow}
\usepackage{array}

\begin{document}

\title{Skeleton-based Gesture Recognition Using Several Fully Connected Layers with Path Signature Features and Temporal Transformer Module}
\author{Chenyang Li, Xin Zhang*, Lufan Liao, Lianwen Jin, Weixin Yang\\
School of Electronic and Information Engineering, South China University of Technology \\
eexinzhang@scut.edu.cn
}


\maketitle
\begin{abstract}
The skeleton based gesture recognition is gaining more popularity due to its wide possible applications. The key issues are how to extract discriminative features and how to design the classification model. In this paper, we first leverage a robust feature descriptor, path signature (PS), and propose three PS features to explicitly represent the spatial and temporal motion characteristics, \emph{i.e.}, spatial PS (S\_PS), temporal PS (T\_PS) and temporal spatial PS (T\_S\_PS). Considering the significance of fine hand movements in the gesture, we propose an "attention on hand" (AOH) principle to define joint pairs for the S\_PS and select single joint for the T\_PS. In addition, the dyadic method is employed to extract the T\_PS and T\_S\_PS features that encode global and local temporal dynamics in the motion. Secondly, without the recurrent strategy, the classification model still faces challenges on temporal variation among different sequences. We propose a new temporal transformer module (TTM) that can match the sequence key frames by learning the temporal shifting parameter for each input. This is a learning-based module that can be included into standard neural network architecture. Finally, we design a multi-stream fully connected layer based network to treat spatial and temporal features separately and fused them together for the final result. We have tested our method on three benchmark gesture datasets, \emph{i.e.}, ChaLearn 2016, ChaLearn 2013 and MSRC-12. Experimental results demonstrate that we achieve the state-of-the-art performance on skeleton-based gesture recognition with high computational efficiency.
\end{abstract}

\section{Introduction}
With the development of intelligent device (\emph{e.g.}, AR, VR and smart-home devices), hand gesture interaction is attracting more attention because of its wide applications for human/computer interaction and communications. Hand gesture recognition is an important and classic problem. Recently, the accurate vision based pose/skeleton estimation gains more popularity due to cost-effective depth sensor (like Microsoft Kinect and Intel RealSense) and reliable real-time body pose estimation development ~\cite{wei2016convolutional}. Comparing with RGB-D sequence based gesture recognition, the skeleton based methods are robust to illumination changes and view variations, and avoid motion ambiguity. In this paper, we focus on the skeleton-based isolated hand gestures recognition, that is, one gesture per one sequence. The key issues are how to extract discriminative spatial temporal features and how to design the classification model.

\begin{figure}[t]
	\begin{center}
		\includegraphics[width=0.45\textwidth]{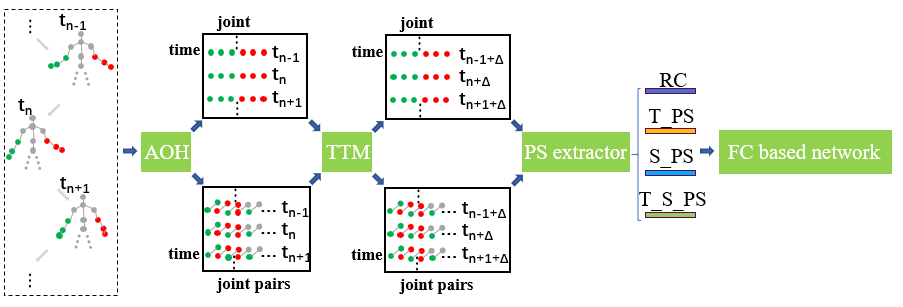}
	\end{center}
	\caption{The flowchart of our algorithm.}
	\label{figure1}
\setlength{\belowcaptionskip}{-1cm}
\end{figure}

Hand gestures can be quite different among various users and application scenario. Since human action and hand gesture are similar in terms of motion representation and problem formulation, here we discuss them together. The gesture recognition framework usually involves the feature description and temporal dependency based classification model. Traditional skeleton-based action recognition approaches involve hand-crafted feature extraction. The joint trajectory covariance matrix~\cite{hussein2013human}, pairwise relative positions~\cite{wang2012mining}, 3D joint histogram~\cite{xia2012view} and Lie group embedding~\cite{vemulapalli2014human} are used to represent the skeleton sequences. Human-crafted features are straightforward but with limited representative abilities, which usually require the domain knowledge. Recently, convolutional neural network (CNN) and its extensions are widely used for feature extraction, like 2D-CNN, 3D-CNN, C3D etc. C3D~\cite{tran2015learning} is a deep 3D convolutional network model based spatial-temporal feature. It is a generic, compact and implicit representation but requires the large-scale training data. Regarding the action temporal dynamics, Fourier temporal pyramid (FTP)~\cite{veeriah2015differential} and hidden Markov model (HMM)~\cite{xia2012view} are used with hand-crafted features. For the deep learning method, different structures of Recurrent Neural Networks (RNN), \emph{e.g.}, hierarchical RNN~\cite{du2015hierarchical}, RNN with regularization~\cite{zhu2016co} differential RNN~\cite{veeriah2015differential}, two-stream RNN~\cite{wang2017modeling} and Long Short-Term Memory (LSTM)~\cite{weng2018deformable}, are popular choice to explore the temporal dependency for recognition. These frameworks have reached state-of-the-art recognition results but the computational complexity may be unacceptable in real-world applications. Hence, we need simple, compact and explicit features to represent global body movements and fine hand motions. Also, the classification model should be simple with temporal dependency.


In this paper, we propose the path-signature feature based hand gesture recognition framework with only few fully connected layers. The flowchart of our algorithm is shown in Fig.~\ref{figure1}. The main contributions are as follows:
\begin{itemize}
	\item We introduce the three different path signature (PS) features , \emph{i.e.}, spatial (S\_PS), temporal (T\_PS) and temporal spatial PS (T\_S\_PS) features, to explicitly characterize spatial configuration and temporal dependency of hand gestures. We also propose an AOH principle to define joint pairs for the S\_PS and select single joint for the T\_PS. In addition, the dyadic method is employed to extract the T\_PS and T\_S\_PS features that encode both global and local temporal dynamics.
	\item We propose the temporal transformer module (TTM) that can actively produce an appropriate temporal transformation for each input sequence. This is a learning-based module that can be included into standard neural network architecture.
	\item We propose an extremely simple multi-streams architecture as the classifier with only several fully connected (FC) layers. Different features have their own channels and the FC layer are used for final fusion.
	\item By only using skeleton data, our method obtains the state-of-the-art results on three major benchmarks, \emph{i.e.}, ChaLearn 2013, ChaLearn 2016 and MSRC 12. Further, our model requires less floating-point multiplication-adds and training memory.
\end{itemize}

\section{Related Work}
\subsection{Gesture recognition}


Skeleton based hand gesture recognition methods are much less than those dealing with the full body skeleton based action recognition \cite{wang2018rgb}. It is limited by the dataset availability and gesture unique property. Hand gesture mainly involves the finger, palm and hand motion, which only has 1-3 joints in the skeleton obtained from depth data. In ~\cite{de2016skeleton}, several skeleton-based features are used together as temporal pyramid, including shape of connected joints, histogram of hand direction and histogram of wrist rotation. In 2D skeleton are superimposed onto original image as dynamic signatures. These features aim to describe the hand motion in detail but the description and representation abilities are limited.



Deep learning have made great process in the area of action recognition. Considering the sequential property, it is natural to apply the RNN, LSTM and their extensions to learn temporal dynamics. In 2017 ChaLearn LAP RGB-D isolated gesture recognition competition ~\cite{wan2017results}, the largest hand gesture recognition contest, most participants (including the winner) used C3D and/or LSTM neural networks. C3D architecture has been widely used in the action recognition for appearance and motion modeling because it is more suitable for spatial-temporal feature extraction than 2D CNN. The model inputs are multi-modalities including RGB, depth, optical flow and/or skeleton. Recently, \cite{weng2018deformable} proposed a deformable pose traversal convolution method based on 1D convolution and LSTM. These recognition networks usually have multiple LSTMs and temporal streams channels and the final result is multi-stream average fusion. In \cite{wang2017modeling}, RNN architecture not only characterizes the temporal dynamics but also considers the spatial configuration in the two-stream architecture. With the proper modeling of skeleton structure and spatial dependency of the action, recognition accuracy increased. In the latest hand gesture recognition research \cite{narayana2018gesture}, by using RGB-D and their flow as inputs, the network has 12 channels representing the large body movement and fine hand motions individually. The fusion channel is a sparsely connected network with one weight per gesture and channel. The RNN/LSTM frameworks deliver the state-of-the-art performance on most action and gesture recognition datasets, indicating the excellent feature and dependency learning capabilities. The only concerns are the architecture complexity, training data requirement and computational efficiency.


\subsection{Path signature feature}
The path signature (PS) was first proposed in~\cite{chen1958integration} in the form of noncommutative formal power series. After that PS was used to solve differential equations driven by rough paths~\cite{lyons1998differential,author9}. Recently, the path signature has been successfully used as a trajectory descriptor and applied to many tasks in the field of machine learning and pattern recognition, such as quantitative finance \cite{gyurko2013extracting,lyons2014feature}, handwriting recognition \cite{lai2017toward,yang2015improved}, writer identification \cite{yang2015chinese,yang2016deepwriterid,liu2017ps}, human action \cite{yang2017leveraging} and hand gesture recognition ~\cite{li2017lpsnet}. ~\cite{yang2017leveraging} is the pioneer work employing the path signature feature for skeleton-based action recognition. All joint pairs and temporal joint evolution are considered as \textit{path} and the corresponding path signatures are computed as features. The concatenation of all path signatures are the input vector for classification. In ~\cite{li2017lpsnet}, the path signature is the firstly used in the gesture recognition by defining the hand trajectory as the path. Path signature can provide the informative representation of sequential data but how to define proper paths and how to deal with their high dimensionality is worthy to be explored.

\section{Overview of Path Signature}
In this section, we will briefly introduce the mathematical definition, geometric interpretation and some properties of path signature (PS), which is mainly referred to \cite{chevyrev2016primer}.

Assume a path $P:[t_1, t_2]\to\mathbb{R}^d$, where $[t_1, t_2]$ is a time interval. The coordinate paths are denoted by $(P_t^1,...,P_t^d)$, where each $P^i:[t_1, t_2]\to\mathbb{R}$ is a real-value path. For an integer $k\ge1$ and the collection of indices $i_1,...,i_k\in\{1,...,d\}$, the \emph{k}-fold iterated integral of the path along indices $i_1,...,i_k$ can be defined as:
\begin{equation}
\begin{split}
S(P)_{t_1,t_2}^{i_1,...,i_k}=\int_{t_1<a_k<t_2}...\int_{t_1<a_1<a_2}dP_{a_1}^{i_1}...dP_{a_k}^{i_k}
\end{split}
\label{eq1}
\end{equation}
where $t_1<a_1<a_2<...<a_k<t_2$.

The signature of path $P$, denoted by $S(P)_{t_1,t_2}$, is the collection (infinite series) of all the iterated integrals of $P$:
\begin{equation}
\begin{split}
S(P)_{t_1,t_2}=&(1,S(P)_{t_1,t_2}^1,S(P)_{t_1,t_2}^2,...,S(P)_{t_1,t_2}^d,\\
&S(P)_{t_1,t_2}^{1,1},...,S(P)_{t_1,t_2}^{1,d},...,S(P)_{t_1,t_2}^{d,d},\\
&...,\\
&S(P)_{t_1,t_2}^{1,...,1},...,S(P)_{t_1,t_2}^{i_1,...,i_k},...,S(P)_{t_1,t_2}^{d,...,d},\\
&...)
\end{split}
\label{eq2}
\end{equation}
The \emph{k}-th level PS is the collection (finite series) of all the \emph{k-fold iterated integral} of path $P$. The \emph{1}-st and \emph{2}-nd level represents path displacement and  path curvature respectively. By increasing k, higher levels of path information can be extracted, but the dimensionality of iterated integrals enlarge rapidly as well. Note that the \emph{0}-th level PS of path $P$ is equal to 1 by convention.

In practice, we often truncate the $S(P)_{t_1,t_2}$ at level $m$ to ensure the dimensionality of the PS feature in a reasonable range. The dimensionality of $S(P)_{t_1,t_2}$ truncated at level $m$ is calculated through $M=d+\cdots+d^m$ (without zeroth term).

The path is considered as the piecewise linear path after sampling. The PS of a discrete path with finite length can be easily calculate based on linear interpolation and Chen's identity \cite{chen1958integration}. For each straight line of a path, the element of its PS can be calculates by:
\begin{equation}
S(P)_{t,t+1}^{i_1,...,i_k}=\frac{1}{k!}\prod_{j=1}^{k}S(P)_{t,t+1}^{i_j}
\label{eq3}
\end{equation}
For the entire path, Chen's identity states that for any $(t_s,t_m,t_u)$ satisfying: $t_s<t_m<t_u$, then,
\begin{equation}
S(P)_{t_s,t_u}^{i_1,...,i_k,...,i_n}=\sum_{k=0}^{n}S(P)_{t_s,t_m}^{i_1,...,i_k}S(P)_{t_m,t_u}^{i_{k+1},i_{k+2},...,i_n}
\label{eq4}
\end{equation}

PS has two excellent properties for path expression. First, PS is the unique representation of a non tree-like path \cite{hambly2010uniqueness}. A tree-like path is a trajectory that retraces itself (such as clapping). For time-sequential data, it's natural and effective to add an extra time dimension into the original path to avoid the tree-like situation. Second, shuffle product identity \cite{lyons2004differential} indicates that the product of two signature of lower level can be expressed as a linear combination of some higher level terms. Hence, adoption of higher level terms of PS actually brings more nonlinear prior knowledge, which reduces the need for the classifier of high complexity. More properties and related details can be found in \cite{chevyrev2016primer}.

We recommend an open-source python library named \emph{iisignature}, which can be easily installed through \emph{pip}.

\begin{figure*}[t]
\begin{center}
\includegraphics[width=\textwidth]{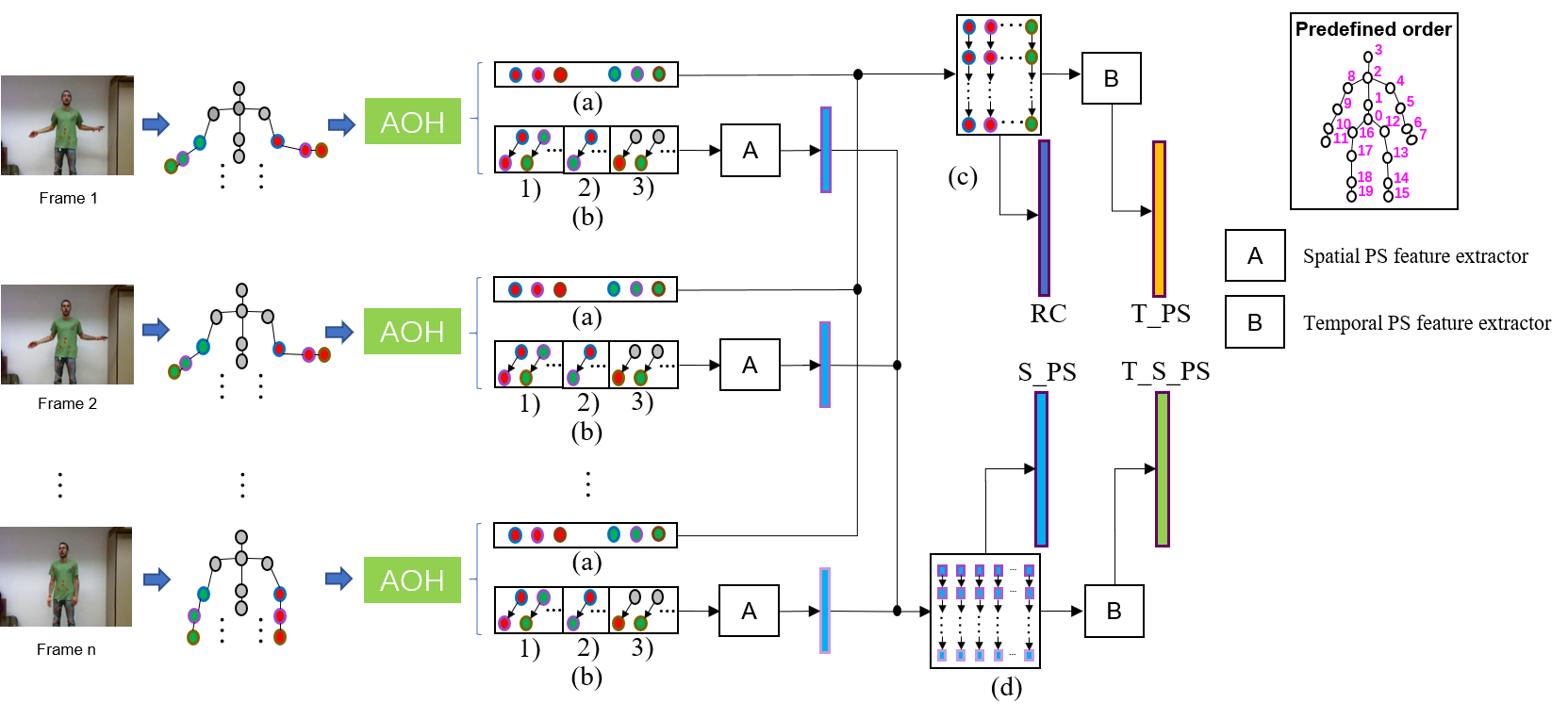}\\
\end{center}
   \caption{The illustration of PS features extraction (T\_PS, S\_PS and T\_S\_PS). (a) and (b) are single joints and joint pairs selected following AOH. (c) and (d) are the temporal paths of single joint and each dimension of S\_PS. Note that TTM (proposed in Section 4.2 and illustrated in Fig.~\ref{figure4}) should be implemented between AOH and PS extractors, but we omit it here for clarity.}
\label{figure2}
\end{figure*}

\section{Approach}
In this section, we first introduce an "attention on hand" (AOH) principle for PS extraction, which deals with global body movements and fine hand motions. Then we propose a novel temporal transformer module (TTM) to alleviate the sequence temporal variation. Finally a multi-stream architecture is presented to fuse different types of features.

\subsection{AOH principle and PS extraction}
\subsubsection{AOH principle}
Before calculating the PS feature, we need to consider about what path to be used and how to design paths efficiently for the recognition. In ~\cite{yang2017leveraging}, the first work leveraging PS features in the human action recognition problem, single joint, joint pair and joint triple are utilized to define paths. They use all $N$ joints and exhaust all the possible pairs and triples (\emph{i.e.}, $C_N^2$ and $C_N^3$), which brings performance improvement but increases dimensionality dramatically. In the context of gesture recognition, we propose the AOH principle to select single joints and joint pairs.

For the \textit{single joint}, only the joint belongs to the hand part (including elbow, wrist and hand 3 joints, \emph{i.e.}, $N_J=3 \cdot 2=6$) are selected, as Fig.~\ref{figure2} (a) shows. For the \textit{joint pair}, three kinds of pairs are considered (as depicted in Fig.~\ref{figure2} (b) 1)-3)). The first kind of \textit{joint pair} is inside the same hand part, describing the geometric characteristics of hand explicitly. The second kind of \textit{joint pair} is from two hand parts, indicating the relative state of two hands. The third kind of \textit{joint pair} is across hand part and body part (upper body joints except hand part), characterizing the related location of hand and body. The selected single joint and joint pairs defined by AOH principle can not only model global hand relative body movements and fine hand motions but also make the PS features more compact.

\subsubsection{Path definition and PS feature extraction}
Based on the selected single joint and joint pairs obtained by AOH principle, we further define one spatial path and two kinds of temporal paths for PS feature extraction. We regard each joint pair as a spatial path for the PS feature extraction. The first type of temporal path is the evolution of each selected single joint along the time, as shown in Fig.~\ref{figure2} (c). Another type of temporal path is the evolution of spatial correlations among joints, as shown in Fig.~\ref{figure2} (d). The summary of three PS features are shown in Table~\ref{table1}.

\textbf{Spatial PS features} The fundamental description of spatial structure is the \emph{d}-dimensional raw coordinates. We concatenate the coordinates of single joints in each frame as a RC vector obeying chronological order as Fig.~\ref{figure2} (c) shows. Further, due to the poor characterization ability of RC and noise interference, we extract PS features over selected joint pairs to explore the spatial relation between joints. The implementations of spatial PS feature extractor (Box A in Fig.~\ref{figure2}) are as follows: i) Select elements that need to be calculated in Eq.\ref{eq1} according to the truncated level $m_S$. ii) Calculate the truncated spatial PS of a joint pair (a ¡°straight line¡± in Eq.\ref{eq3}) by Eq.\ref{eq1} and Eq.\ref{eq3} (The start and end points are defined according to the predefined order in Fig.~\ref{figure2}). iii) Finally concatenate the truncated spatial PS of all input joint pairs as the spatial PS (S\_PS) feature.

\textbf{Dyadic temporal PS features} The dyadic method with PS was firstly used in \cite{yang2016rotation} for the writer ID identification. Since the gesture always contain global and local variation, we employ the dyadic method for the temporal PS feature extraction. The dyadic method divides the entire path into dyadic pieces and set up a hierarchical representation of the path, which can extract both the global and local feature of entire path, and reduce the feature dimensionality as well. If the dyadic level is $L_{D}$, then an entire path can be divided into $2^{(L_{D}+1)}-1$ subpaths.

To characterize the temporal dynamic of single joint, the evolution of each single joint is treated as an entire temporal path , as shown in Fig.~\ref{figure2} (c). An extra monotone time dimension is added to ensure the path uniqueness (\emph{i.e.}, to avoid tree-like path as discussed in Section 3).

To further explore kinematic constraints of the joint pairs, the evolution of each dimension in the S\_PS of every frame also can be regarded as another kind of entire temporal path, as Fig.~\ref{figure2} (d) shows. As a result, we acquire a series of 1D paths. However, the signature of a 1D path is just the increments to a certain power, which can be easily get from Eq.~\ref{eq3}. To alleviate this problem, we use the lead-lag transformation as ~\cite{yang2017leveraging} does over the 1D path to enrich the temporal contextual information.

The implementations of temporal PS feature extractor (Box B in Fig.~\ref{figure2}) are as follows: i) Select elements that need to be calculated in Eq.\ref{eq1} according to the truncated level $m_T$ or $m_{T\_S}$. ii) Every subpath generated by dyadic method is an ¡°entire path¡± (consist of several straight lines) in Eq.~\ref{eq4}. According to Eq.~\ref{eq1}, Eq.~\ref{eq3} and Eq.~\ref{eq4}, the truncated temporal PS of a subpath can be calculated. iii) Concatenate the truncated temporal PS of all subpaths as temporal PS (T\_PS) or temporal spatial PS (T\_S\_PS) feature.

\makeatletter
\def\hlinew#1{
  \noalign{\ifnum0=`}\fi\hrule \@height #1 \futurelet
   \reserved@a\@xhline}
\makeatother

\begin{table}[h]
\begin{tabularx}{\columnwidth}{c|X}
\hlinew{1pt}
\textbf{Feature types}                                                            & \multicolumn{1}{c}{\textbf{Feature description}}                                                                                                  \\ \hlinew{1pt}
\begin{tabular}[c]{@{}c@{}}Raw coordinates \\ (\textbf{RC})\end{tabular}           & The \emph{d}-dimensional coordinates of $N_J$ single joints.       \\ \hline
\begin{tabular}[c]{@{}c@{}}Spatial PS \\ (\textbf{S\_PS})\end{tabular}             & The PS over each predefined joint pair truncated at level $m_S$.                                                        \\ \hline
\begin{tabular}[c]{@{}c@{}}Temporal PS \\ (\textbf{T\_PS})\end{tabular}            & The PS over the temporal evolution of each single joint truncated at level $m_T$.                                                  \\ \hline
\begin{tabular}[c]{@{}c@{}}Temporal Spatial PS \\ (\textbf{T\_S\_PS})\end{tabular} & The PS over the temporal evolution of each dimension of S\_PS truncated at level $m_{T\_S}$. \\ \hlinew{1pt}
\end{tabularx}
\caption{The proposed feature for skeletal hand gesture recognition.}
\label{table1}
\end{table}

\subsection{Temporal Transformer Module (TTM)}
\subsubsection{Motivation}
Although deep neural network achieves break through in the sequential recognition task, it's still limited by the lack of ability to be temporally invariant to the input sequence in a computationally and parameter efficient manner. In the context of action recognition, the time-stamps of keyframes are variant among different clips, which makes the model difficult to catch the key information along time dimension.

\begin{figure}[t]
\begin{center}
\includegraphics[width=0.45\textwidth]{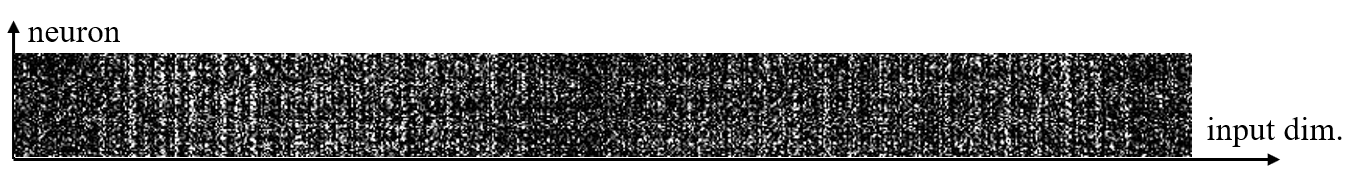}
\end{center}
    \caption{The visualization of the weight matrix of first fully connected layer in a trained one-stream network.}
\label{figure3}
\end{figure}

There are mainly two existing types of methods to solve this problem: structure driven method and data driven method. The structure driven method mostly use LSTM to model the temporal contextual dependence of sequence data. The data driven method is to provide more diverse samples by temporal shift data enhancement. However, LSTM model requires large training data and unnegligible training cost. If we used a simple network like FC layer as the classifier, the temporal consistency is also learned as part of features, which is the unwanted result. For example, we visualize the weight matrix of the first FC layer of a trained one-stream network (will be introduced in the following), which takes RC as input, as shown in Fig.~\ref{figure3}. The x-axis denotes the input dimensionality (obey chronological order), and the y-axis denotes the neuron number (64 in the first FC layer). The brighter position means the corresponding weight is larger, that is, this connection is more important. The FC layer pay more attention on several time stamp, indicating the position of key frames. If there is the temporal variation between the model and testing sequence, the recognition result is worse. Even if the training data is augmented by temporal shift, the model capacity is too small to fit any arbitrary temporal situation.

Recent work \cite{cao2017egocentric} proposed a spatiotemporal transform method to deal with the spatiotemporal variation, but their method is for RGB-D video. As mentioned in Section 4.1, we have employed the temporal PS features to represent temporal dynamics within the sequence. This is what exactly \cite{yang2017leveraging} has done. The inter-sequence temporal difference might be alleviated by the temporal transformation. To this end, we design a differentiable module called temporal transformer module (TTM). This module can actively transform the input data temporally and finally adjust the key frame to the best time stamp for the network.

\subsubsection{Proposed TTM}
Inspired by STN \cite{jaderberg2015spatial}, TTM is a differentiable module that applies a temporal transform to RC. The TTM contains two steps: Localization network (LN) and temporal shifting.

\begin{figure*}[t]
\begin{center}
\includegraphics[width=6in]{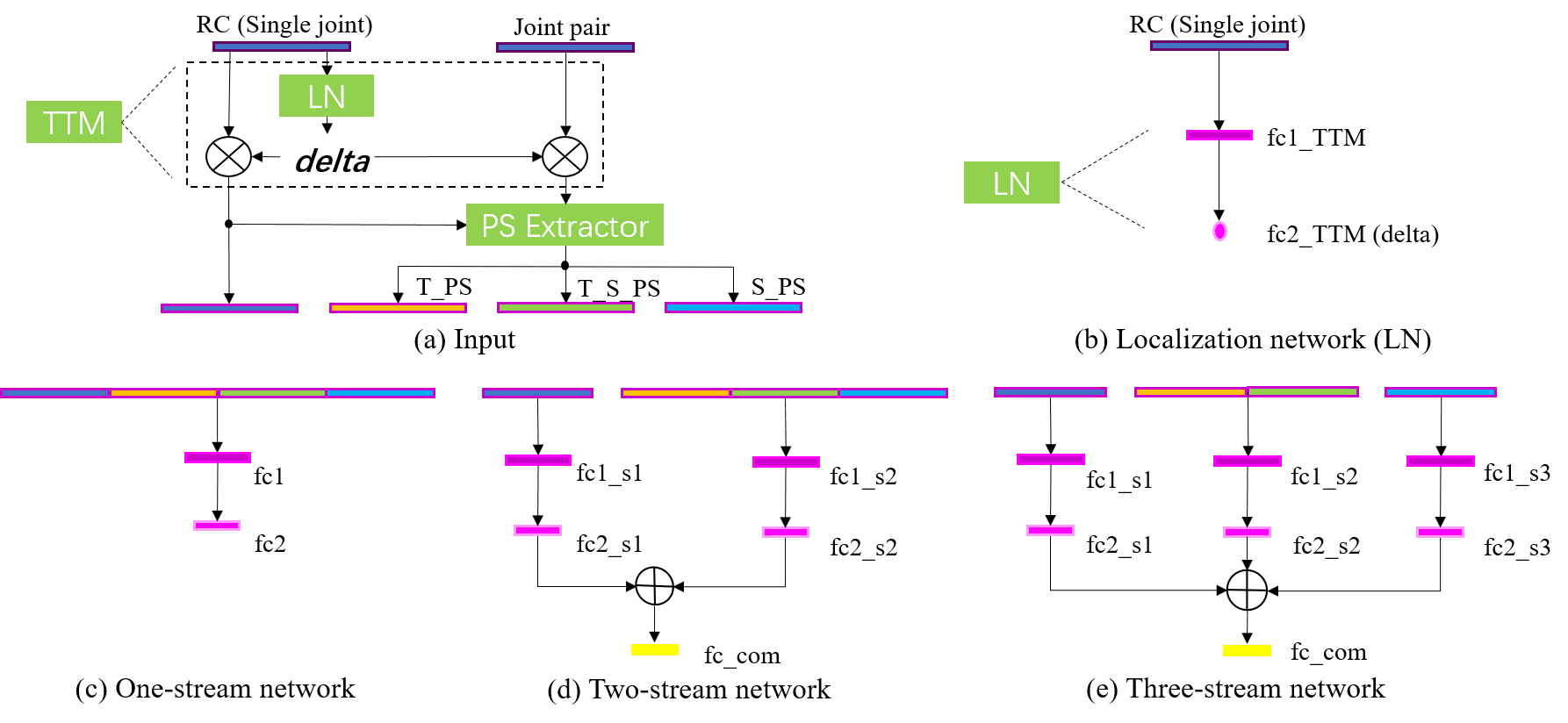}\\
\end{center}
   \caption{Three different types of network architectures: one-stream network (1s\_net), two-stream network (2s\_net) and three-stream network (3s\_net). TTM is the temporal transformer module. LN is the localization network that generates the transformation parameters $\Delta$. $\bigotimes$ and $\bigoplus$ denote temporal shift and weighted sum. The fc2 in each stream denotes a FC layer with a softmax activation function.}
\label{figure4}
\end{figure*}

Firstly, we use LN to learn the temporal transform factor delta ($\Delta$), as shown in Fig.\ref{figure4} (b). It takes the input vector $I \in \mathbb{R}^{D_{RC}}$ ($D_{RC}$ denotes the dimensionality of RC) and output $\Delta$ as shown in Fig.~\ref{figure4} (b), \emph{i.e.}, $\Delta=f_{LN}(I)$. Note that the network function $f_{LN}()$ can take any form, such as FC layer or 1D convolution layer, but should finally regress to one neuron.

Secondly, the input vector $I$ is reshaped as a matrix $V^i \in \mathbb{R}^{d \cdot N_J \times F}$, where $N_J$ is the number of single joints (\emph{i.e.}, 6 in Fig.~\ref{figure2}) and $F$ denotes the frame number. Each column of $V^i$ is a vector $v_x^i$ which consists of the coordinates of single joints in the same frame, $x \in [1, F]$. And if we denote the matrix after shifting as $V^o$, then each column of it can be calculate by:
\begin{equation}
v_x^o=f_I(x-\Delta, V^i)=(1 - \alpha) \cdot v_{\left \lfloor x-\Delta \right \rfloor}^i + \alpha \cdot v_{\left \lceil x-\Delta \right \rceil}^i
\label{eq5}
\end{equation}
Here, because $\Delta$ is a decimal, we use linear interpolation function $f_I()$ to generate the $V^o$. $\alpha$ is calculated by $(x-\Delta) - \left \lfloor x-\Delta \right \rfloor$. Note that $x-\Delta$ is clip by value [1, F]. Eventually, $V^o$ is reshaped back as a output vector $O$.

\subsection{Multi-stream Architecture}
As discussed in Section 4.1, we define spatial and temporal path based on AOH principle. Corresponding PS features encode the spatial and temporal information of the action. Previous work \cite{yang2017leveraging,li2017lpsnet} concatenates these features together as whole and one classification model is employed, \emph{e.g}, FC layers. We believe temporal and spatial features should be treated separately because one represents the joint evolution and the other describes the body configuration. The final result is fusion of these multiple channels.

So we utilize the multi-stream architecture to process different kinds of information separately. As a result, we design three kinds of network architectures, one-stream network (1s\_net), two-stream network (2s\_net) and three-stream network (3s\_net), as shown in Fig.~\ref{figure4}(c)-(e) . The 1s\_net directly concatenates all the features as one input vector and feed it to a 2-fc-layer network, similar to \cite{yang2017leveraging}. The 2s\_net has two inputs, RC and PS, representing basic information and extracted compact information. As defined in Table \ref{table1}, the T\_PS and T\_S\_PS represent temporal information and S\_PS is spatial feature. Hence, the 3s\_net has three streams with two FC layers separately. The final fusion result is obtained through a FC layer as the weighted summation.

\section{Experimental Results and Discussion}
\subsection{Datasets}
\textbf{ChaLearn 2013 dataset:} It is the ChaLearn 2013 Multi-model gesture dataset \cite{escalera2013multi}, which contains 23 hours of Kinect data with 27 persons performing 20 Italian gestures. This dataset provides RGB, depth, foreground segmentation and Kinect skeletons. Here, we only use skeletons for gesture recognition as done in the literature \cite{wang2017modeling}.

\noindent\textbf{ChaLearn 2016 dataset:} ChaLearn Isolated dataset \cite{wan2016chalearn}, the largest gesture recognition dataset consisting of RGB and depth videos, includes 35,878 training, 5,784 validation and 6,271 test videos for totally 249 gestures. We use Openpose \cite{wei2016convolutional} to estimate the skeleton joints in all videos as \cite{lin2018large} did. It can be downloaded from http://www.hcii-lab.net/data/.

\noindent\textbf{MSRC-12 dataset:} MSRC-12 gesture dataset \cite{fothergill2012instructing} includes 6 iconic and 6 metaphoric gestures performed by 30 people. We use 6 iconic gestures from the dataset that amounts to 3,034 instances and employ 5-fold leave-person-out cross-validation as in \cite{jung2014enhanced}.

\subsection{Data Preprocessing and Network Setting}
We first normalize skeletons by subtracting the central joint, which is the average position of all joints in a video clip. Then all coordinates are further normalized to the range of [-1, 1] over the entire video clip. Finally, we sample all videos clips to 39 frames by linear interpolation or uniform sampling. The data enhancement methods we use are three-fold. The first one is temporal augmentation by randomly temporal shift the frame in range of [-5, 5]. The second one is adding Gaussian noise with a standard deviation of 0.001 to joints coordinates. The last one is rotating coordinates along x, y, z three axes in range of [$-\pi/36$, $\pi/36$], [$-\pi/18$, $\pi/18$] and [$-\pi/36$, $\pi/36$].

For ChaLearn 2013 and MSRC-12 two datasets, we set the neuron number of each 2-fc-layer stream to 64 and C (64\_fc-C\_fc), where C is the gesture class number. For the largest dataset ChaLearn 2016, we use 256\_fc-C\_fc. We adopt 64\_fc-1\_fc architecture for $f_{LN}()$. DropOut \cite{hinton2012improving} layer is added after the first FC layer of each stream to avoid over fitting. The mini-batch size and dropout rate are set to 56 and 0.5. We use the method of stochastic gradient descent with a momentum value equal to 0.7.  The learning rate updates in accordance to $\alpha(n)=\alpha(0)\cdot exp(-\lambda n)$, where $n$ is the cumulative mini-batch number. $\alpha(0)$ and $\lambda$ are set to 0.01 and 0.001. The $L_D$ for T\_PS and T\_S\_PS calculation are set to 3 and 2. The lead-lag dimensionality is set to 2.

\subsection{Ablation Study}
We do some ablation experiments on the ChaLearn 2013 dataset to explore the truncated level of PS and examine the effectiveness of PS, TTM and multi-stream architecture.
\subsubsection{Investigation of the truncated level of PS}
We utilize the one-stream network without TTM (1s\_net\_w/o TTM) to explore the contributions of different truncated level PS. The validation  accuracy rate is 77.84\% with only RC. The accuracy rate after adding different PS are shown in Table~\ref{table2}. It is noted that the T\_S\_PS is calculated from S\_PS truncated at level 2 (The abbreviation can be referred to Table~\ref{table1}). The performance improves after adding any type of PS truncated at any level, which indicates the effectiveness of PS. The last column of each row illustrates that all types of PS feature are complementary.

It is worthy to note that the contributions trend to be negligible and even vanish when truncated level is greater than a certain value. There is a trade-off between validation performance and feature dimensionality. As a result, we choose to set $m_T$, $m_S$, $m_{T\_S}$ as 4, 2, 3.

\begin{table}[h]
\setlength{\tabcolsep}{1.35mm}{
\begin{tabular}{c|c|c|c|c}
\hlinew{1pt}
\textbf{PS truncated level} & \textbf{+T\_PS} & \textbf{+S\_PS} & \textbf{+T\_S\_PS} & \textbf{+AllPS} \\ \hlinew{1pt}
1                           & 77.98          & 82.30          & 85.35            & \textbf{85.50}               \\ \hline
2                           & 81.28          & 87.62          & 87.80            & \textbf{88.76}  \\ \hline
3                           & 84.42          & 88.12          & 88.81            & \textbf{89.17}  \\ \hline
4                           & 85.27          & 88.06          & 88.09            & \textbf{89.20}               \\ \hlinew{1pt}
\end{tabular}}
\caption{The ablation study of PS features on ChaLearn 2013. The truncated level of PS can be referred to Section 3.}
\label{table2}
\end{table}

\subsubsection{Investigation of the TTM}
We use the 1s\_net with RC as input to examine the effectiveness of TTM. Results are presented in Table~\ref{table3}. 1s\_net\_w/o TTM can be roughly regarded as the method proposed by \cite{yang2017leveraging}.

Firstly, we use temporal augmentation to test whether the data driven methods can make the improvement. We shift the samples in range of [-5, 5] to provide more diverse samples. However, it doesn't improve the performance. Hence, directly data driven methods cannot work well.

Then we directly add TTM to the 1s\_net (1s\_net\_TTM), and the result improves from 77.84\% to 80.55\%. The temporal transformation parameter learned by TTM can fit the key moment of an action and the active part of the FC layer.

At last we add the same temporal augmentation  for 1s\_net\_TTM, and the accuracy rate increases from 80.55\% to 81.33\%. This attractive observation indicates that TTM makes good use of the diverse samples provided by temporal enhancement. In other words, the network is more adaptive after adding TTM.

\begin{table}[h]
\setlength{\tabcolsep}{2.8mm}{
\begin{tabular}{c|c}
\hlinew{1pt}
\textbf{Components}   & \textbf{Accuracy rate (\%)} \\ \hlinew{1pt}
1s\_net\_w/o TTM             & 77.84                       \\ \hline
1s\_net\_w/o TTM + Temp. Enh.      & 77.46                       \\ \hline
1s\_net\_TTM          & 80.55                       \\ \hline
1s\_net\_TTM + Temp. Enh. & 81.33                       \\ \hlinew{1pt}
\end{tabular}}
\caption{The ablation study of TTM on ChaLearn 2013. Temporal Enhancement is abbreviated to Temp. Enh.}
\label{table3}
\end{table}

\subsubsection{Investigation of different network architectures}
For the estimation of different architecture, we use network without TTM, with RC and all PS selected above as input, which can be roughly  regarded as the architecture utilized by \cite{yang2017leveraging}. As shown in Table~\ref{table4}, the performance improves clearly, which indicates that the multi-stream architectures allow each stream to dig deeply into one type of feature and finally provide more discriminative information.

\begin{table}[h]
\setlength{\tabcolsep}{3.75mm}{
\begin{tabular}{c|c|c|c}
\hlinew{1pt}
\textbf{Components}        & 1s\_net & 2s\_net & 3s\_net \\ \hlinew{1pt}
\textbf{Accuracy rate (\%)} & 89.43   & 89.63   & 90.19   \\ \hlinew{1pt}
\end{tabular}}
\caption{The ablation study of network architectures.}
\label{table4}
\end{table}

\subsection{Comparison with the State-of-the-arts}
In this subsection, we use the best parameter setting and network structure getting from our ablation study. We also do all the data augmentation methods mentioned above for our network.
\subsubsection{ChaLearn 2013}
The results on the ChaLearn 2013 dataset are shown in Table~\ref{table5}. Currently, methods achieving the best performance are mainly benefited from powerful characterization ability of CNN and LSTM models. \cite{du2015skeleton} organizes the raw coordinates as the spatial temporal feature maps then feeds it to the hierarchical spatial-temporal adaptive filter banks CNN architecture. \cite{wang2017modeling} propose a two-stream LSTM to model temporal dynamics and spatial configurations separately. Compared with these method, our FC based network achieves the best results with less multiplication-adds operation as Table~\ref{table6} shows. Note that the AOH principle dramatically reduces the Multiplication-Adds compared with the method without AOH \cite{yang2017leveraging}.

\begin{table}[h]
\setlength{\tabcolsep}{3mm}{
\begin{tabular}{c|c}
\hlinew{1pt}
\textbf{Method}  & \textbf{\begin{tabular}[c]{@{}c@{}}Accuracy \\ rate (\%)\end{tabular}} \\ \hlinew{1pt}
HiVideoDarwin \citeauthor{wang2015hierarchical}  & 74.90                       \\
VideoDarwin \citeauthor{fernando2015modeling}    & 75.30                       \\
D-LSDA \citeauthor{su2018discriminative} & 76.80                       \\
CNN for Skeleton \citeauthor{du2015skeleton} & 91.20                        \\
Two-stream LSTM \citeauthor{wang2017modeling}   & 91.70                        \\ \hline
3s\_net\_TTM     & \textbf{92.08}              \\ \hlinew{1pt}
\end{tabular}}
\caption{Comparison of methods on the ChaLearn 2013.}
\label{table5}
\end{table}

\begin{table}[h]
\setlength{\tabcolsep}{0.1mm}{
\begin{tabular}{c|c|c|c|c}
\hlinew{1pt}
\textbf{Method}                                                        & CNN$^{[1]}$   & 2s\_LSTM$^{[2]}$ & 3s\_net w/o AOH & 3s\_net   \\ \hlinew{1pt}
\textbf{\begin{tabular}[c]{@{}c@{}}Mult-adds\\ (Million)\end{tabular}} & 11.54 & 358.34   & 14.89$^{1}$+15.06     & \textbf{2.69$^{1}$+2.00} \\ \hlinew{1pt}
\end{tabular}}
\caption{Comparison of the Multiplication-Adds. $^{[1]}$: \citeauthor{du2015skeleton} $^{[2]}$: \citeauthor{wang2017modeling} $^{1}$: PS calculation. }
\label{table6}
\end{table}

\subsubsection{ChaLearn 2016}
The results on ChaLearn 2016 are summarized in Table~\ref{table7}. Our model outperforms the skeleton based method \cite{lin2018large} by around 4.5\%. We also notice that the accuracies of skeleton based methods are inferior to video frame based models. The reasons are mainly two-fold. Firstly, the precision of OpenPose is affected by the drastic background and illumination changes. Secondly, a lot of gesture classes requires recognizing the static hand gesture instead of dynamic hand motion. The recognition performance on these classes mainly depends on the estimation precision of hand joints. It is worth noting that our model is the simplest one. We argue that the performance will improve if more accurate joints locations are provided.

\begin{table}[h]
\setlength{\tabcolsep}{0.3mm}{
\begin{tabular}{c|c|c|c|c|c|c}
\hlinew{1pt}
\multirow{2}{*}{\textbf{Method}} & \multirow{2}{*}{\textbf{\begin{tabular}[c]{@{}c@{}}Test acc.\\ (\%) \end{tabular}}} & \multicolumn{4}{c|}{\textbf{Modality}}            & \multirow{2}{*}{\textbf{Model}} \\ \cline{3-6}
                                 &                                                                               & \textbf{R} & \textbf{D} & \textbf{O} & \textbf{S} &                        \\ \hlinew{1pt}
AMRL$^{[1]}$                             & 65.59                                                                        & $\surd$          & $\surd$          &           &           & 8*CNN+4ConvLSTM        \\ \cline{7-7}
ASU$^{[2]}$                              & 67.71                                                                        & $\surd$          & $\surd$          & $\surd$          &           & 4*C3D+2*TSN+1*SVM      \\ \cline{7-7}
FOANet$^{[3]}$                           & \textbf{82.07}                                                                         & $\surd$          & $\surd$          & $\surd$          &           & 12*CNN                 \\ \hline
SkeLSTM$^{[4]}$                          & 35.39                                                                         &           &           &           & $\surd$          & 1*LSTM                 \\ \hline
3s\_net\_TTM                             & \textbf{39.95}                                                                             &           &           &           & $\surd$          & 3*FC                   \\ \hlinew{1pt}
\end{tabular}}
\caption{Comparison on ChaLearn 2016 dataset. $^{[1]}$: \citeauthor{wang2017large} $^{[2]}$: \citeauthor{miao2017multimodal} $^{[3]}$: \citeauthor{narayana2018gesture} $^{[4]}$: \citeauthor{lin2018large} RGB, depth, optical flow and skeleton are abbreviated as R, D, O and S.}
\label{table7}
\end{table}

\subsubsection{MSRC-12}
\cite{wang2016action} proposed Joint Trajectory Maps (JIM), which are a set of 2D images that encode spatiotemporal information carried by 3D skeleton sequences in three orthogonal planes. In \cite{jung2014enhanced}, a novel framework called Enhanced Sequence Matching (ESM) is leveraged to align and compare action sequences based on  a set of elementary Moving Poses (eMP). \cite{garcia2017transition} proposed "transition forests", an ensemble of randomized tree classifiers that learnt both static pose information and temporal transitions. All these methods show the importance of spatio-temporal information modelling. Our method extracts spatial, temporal and joint spatial-temporal features and achieves the state-of-the-art accuracy of 99.01\%, as shown in Table \ref{table8}.

\begin{table}[h]
\setlength{\tabcolsep}{3.8mm}{
\begin{tabular}{c|c}
\hlinew{1pt}
\textbf{Method}  & \textbf{\begin{tabular}[c]{@{}c@{}}Accuracy \\ rate (\%)\end{tabular}} \\ \hlinew{1pt}
JTM \citeauthor{wang2016action} & 93.12                       \\
DFM \citeauthor{lehrmann2014efficient} & 94.04                       \\
ESM \citeauthor{jung2014enhanced}  & 96.76                       \\
RJP \citeauthor{garcia2017transition}    & 97.54                       \\
MP \citeauthor{garcia2017transition} & 98.25                        \\ \hline
3s\_net\_TTM     & \textbf{99.01}              \\ \hlinew{1pt}
\end{tabular}}
\caption{Comparison of methods on the MSRC-12 dataset.}
\label{table8}
\end{table}

\section{Conclusion}
In this paper, we first leverage S\_PS, T\_PS and T\_S\_PS three PS features to explicitly represent the spatial and temporal motion characteristics. In the path definition, we propose the AOH principle to select single joint and joint pairs, which ensures the feature robust and compact. Furthermore, the dyadic method employed to extract the T\_PS and T\_S\_PS features that encode global and local temporal dynamics with less dimensionality. Secondly, we propose a differentiable module TTM to match the sequence key frames by learning the temporal shifting parameter for each input. Finally, we design a multi-stream FC layer based network to treat spatial and temporal features separately. The ablation study has shown the effective of every contribution. We have achieved the best result on skeleton-based gesture recognition with high computational efficiency on three benchmarks. We will explore the possible combination of the attention scheme and PS features.

\section{Acknowledgments}
This work is supported by GD-NSF (2016A010101014, 2017A030312006, 2018A030313295), the Science and Technology Program of Guangzhou (2018-1002-SF-0561), the National Key Research and Development Program of China (2016YFB1001405), the MSRA Research Collaboration Funds (FY18-Research-Theme-022), Fundamental Research Funds for Central Universities of China (2017MS050). We thank Prof. Terry Lyons from University of Oxford and Dr. Hao Ni from UCL for their great help. We thank anonymous reviewers for their careful reading and insightful comments.

{\small
\bibliographystyle{aaai}
\bibliography{egbib}
}

\end{document}